\documentclass{article}

     \usepackage[nonatbib,preprint]{neurips_2019}

\usepackage[utf8]{inputenc} % allow utf-8 input
\usepackage[T1]{fontenc}    % use 8-bit T1 fonts
\usepackage{hyperref}       % hyperlinks
\usepackage{url}            % simple URL typesetting
\usepackage{booktabs}       % professional-quality tables
\usepackage{amsfonts}       % blackboard math symbols
\usepackage{nicefrac}       % compact symbols for 1/2, etc.
\usepackage{microtype}      % microtypography
\usepackage{acronym}
\usepackage{comment}
\usepackage{arydshln}
\usepackage{graphicx}
\usepackage{amsmath}
\usepackage{amssymb}

\newacro{cnn}[\textsc{Cnn}]{Convolutional neural network}

\begin{document}

\title{Uncertainty Estimations by Softplus normalization in Bayesian Convolutional Neural Networks with Variational Inference}

\author{Kumar Shridhar\thanks{Authors contributed equally.} \\
  Department of Computer Science \\
  TU Kaiserslautern\\
  MindGarage \\
  %% Address \\
  \texttt{k\_shridhar16@cs.uni-kl.de} \\
  \and 
  Felix Laumann\footnotemark[1] \\
  Department of Mathematics \\
  Imperial College London\\
  NeuralSpace \\
  \texttt{f.laumann18@imperial.ac.uk} \\[1ex]
  \and 
  Marcus Liwicki \\
  Department of Computer Science \\
  Lule\aa \ University of Technology \\
  MindGarage \\
  \texttt{marcus.liwicki@ltu.se}
}

\maketitle

\begin{abstract}
We introduce a novel uncertainty estimation for classification tasks for Bayesian convolutional neural networks with variational inference. By normalizing the output of a Softplus function in the final layer, we estimate aleatoric and epistemic uncertainty in a coherent manner. The intractable posterior probability distributions over weights are inferred by \textit{Bayes by Backprop}. Firstly, we demonstrate how this reliable variational inference method can serve as a fundamental construct for various network architectures. On multiple datasets in supervised learning settings (MNIST, CIFAR-10, CIFAR-100), this variational inference method achieves performances equivalent to frequentist inference in identical architectures, while the two desiderata, a measure for uncertainty and regularization are incorporated naturally. Secondly, we examine how our proposed measure for aleatoric and epistemic uncertainties is derived and validate it on the aforementioned datasets.

\begin{comment}
Recent work in computer vision has focused on improving model accuracy without assessing the uncertainty measure in the neural network predictions. This is mainly because of a lack of probabilistic interpretation in a point estimate network making them unsuitable for the job. In this paper, we introduce a novel uncertainty estimation using Bayesian Convolutional Neural Networks with Variational Inference that performs two sequential convolutions to establish a relationship between the variance and the mean of a multinomial random variable. (add something)  replaces the Softmax function with a normalization of Softplus outputs. The intractable posterior probability distributions over weights are inferred by {Bayes by Backprop}.  Firstly, we demonstrate how this reliable variational inference method can serve as a fundamental construct for various network architectures. On multiple datasets in supervised learning settings (MNIST, CIFAR-10, CIFAR-100), this variational inference method achieves performances equivalent to frequentist inference in identical architectures, while the two desiderata, a measure for uncertainty and regularization are incorporated naturally. Secondly, we examine how our proposed measure for aleatoric and epistemic uncertainties is derived and validate it on the aforementioned datasets.
\end{comment}
\end{abstract}

\section{Introduction}
\begin{figure}[t] 
\centering
\begin{minipage}{.35\textwidth}
\centering
\includegraphics[width=\linewidth]{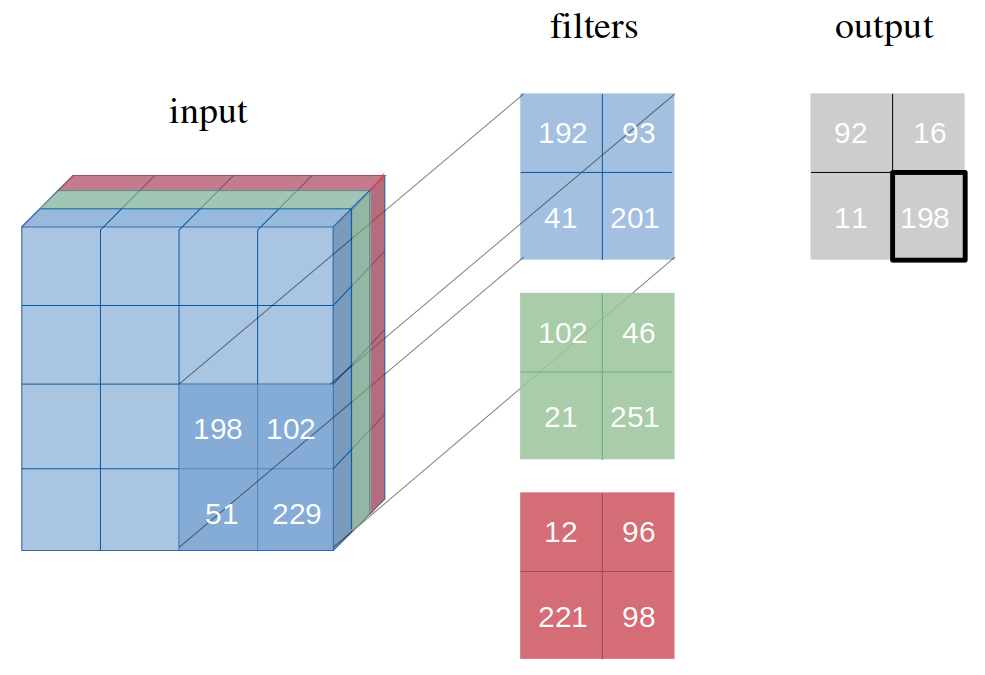}
\end{minipage}
\begin{minipage}{.35\textwidth}
\centering
\includegraphics[width=\linewidth]{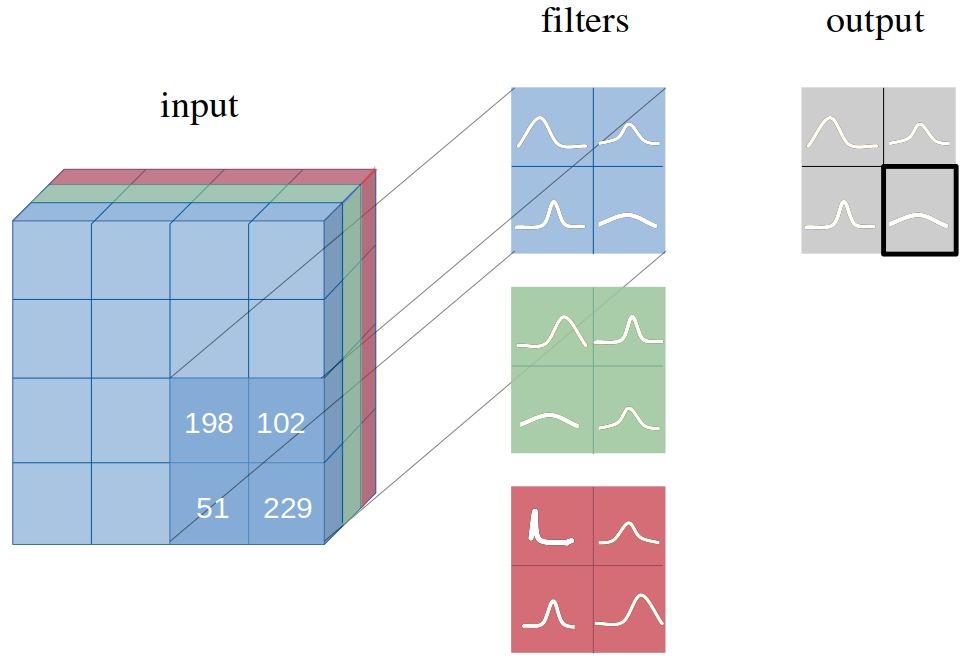}
\end{minipage}
\caption{Input image with exemplary pixel values, filters, and corresponding output with point-estimates (left) and probability distributions (right) over weights.}
\label{fig:filter_scalar}
\end{figure}

\acp{cnn} excel at tasks in the realm of image classification (e.g. \cite{he2016deep,simonyan2014very,krizhevsky2012imagenet}). However, from a probability theory perspective, it is unjustifiable to use single point-estimates as weights to base any classification on. \acp{cnn} with frequentist inference require substantial amounts of data examples to train on and are prone to overfitting on datasets with few examples per class.
\newline In this work, we apply Bayesian methods to \acp{cnn} in order to add a measure for uncertainty and regularization in their predictions, respectively their training. This approach allows the network to express uncertainty via its parameters in form of probability distributions (see Figure \ref{fig:filter_scalar}). At the same time, by using a prior probability distribution to integrate out the parameters, we compute the average across many models during training, which gives a regularization effect to the network, thus preventing overfitting.
\newline We build our Bayesian \ac{cnn} upon \textit{Bayes by Backprop} \cite{graves2011practical,blundell2015weight}, and approximate the intractable true posterior probability distributions $p(w|\mathcal{D})$ with variational probability distributions $q_{\theta}(w|\mathcal{D})$, which comprise the properties of Gaussian distributions $\mu \in \mathbb{R}^d$ and $\sigma \in \mathbb{R}^d$, denoted $\mathcal{N}(\theta|\mu, \sigma^2)$, where $d$ is the total number of parameters defining a probability distribution. The shape of these Gaussian variational posterior probability distributions, determined by their variance $\sigma^2$, expresses an uncertainty estimation of every model parameter. The main contributions of our work are as follows: 
\begin{enumerate}
    \item We present how \textit{Bayes by Backprop} can be efficiently applied to \acp{cnn}. We therefore introduce the idea of applying two convolutional operations, one for the mean and one for the variance.
    \item We empirically show how this generic and reliable variational inference method for Bayesian \acp{cnn} can be applied to various \ac{cnn} architectures without any limitations on their performances, but with intrinsic regularization effects. We compare the performances of these Bayesian \acp{cnn} to \acp{cnn} which use single point-estimates as weights, i.e. which are trained by frequentist inference.
    \item We explain and implement Softplus normalization by means of examining how to estimate aleatoric and epistemic uncertainties without employing an additional Softmax function in the output layer \cite{kwon2018uncertainty} which brings an inconsistency of activation functions into the model.
\end{enumerate} 
This paper is structured as subsequently outlined: after we have introduced our work here, we secondly review Bayesian neural networks with variational inference, including previous works, an explanation of \textit{Bayes by Backprop} and its implementation in \ac{cnn}. Thirdly, we examine aleatoric and epistemic uncertainty estimations with an outline of previous works and how our proposed method directly connects to those. Fourthly, we present our results and findings through experimental evaluation of the proposed method on various architectures and datasets before we finally conclude our work. 

%\begin{comment}
%
%
%\end{comment}

\section{Bayesian convolutional neural networks with variational inference}
Recently, the uncertainty afforded by \textit{Bayes by Backprop} trained neural networks has been used successfully to train feedforward neural networks in both supervised and reinforcement learning environments \cite{blundell2015weight, lipton2016efficient, houthooft2016curiosity}, for training recurrent neural networks \cite{fortunato2017bayesian}, and for \acp{cnn} \cite{shridhar2018BayesianComprehensive, neklyudov2018variance}. Here, we review this method for \acp{cnn} to construct a common foundation on which be build on in section \ref{uncertainty}.
\subsection{Related work}
Applying Bayesian methods to neural networks has been studied in the past with various approximation methods for the intractable true posterior probability distribution $p(w|\mathcal{D})$. \cite{buntine1991bayesian} started to propose various \textit{maximum-a-posteriori} (MAP) schemes for neural networks. They were also the first who suggested second order derivatives in the prior probability distribution $p(w)$ to encourage smoothness of the resulting approximate posterior probability distribution. In subsequent work by \cite{hinton1993keeping}, the first variational methods were proposed which naturally served as a regularizer in neural networks. \cite{hochreiter1995simplifying} suggest taking an information theory perspective into account and utilizing a minimum description length (MDL) loss. This penalises non-robust weights by means of an approximate penalty based upon perturbations of the weights on the outputs. \cite{denker1991transforming} and \cite{mackay1995probable} investigated the posterior probability distributions of neural networks by using Laplace approximations. As a response to the limitations of Laplace approximations, \cite{neal2012bayesian} investigated the use of hybrid Monte Carlo for training neural networks, although it has so far been difficult to apply these to the large sizes of neural networks built in modern applications. More recently, \cite{graves2011practical} derived a variational inference scheme for neural networks and \cite{blundell2015weight} extended this with an update for the variance that is unbiased and simpler to compute. \cite{graves2016stochastic} derives a similar algorithm in the case of a mixture posterior probability distribution. 
\newline Several authors have derived how Dropout \cite{srivastava2014dropout} and Gaussian Dropout \cite{wang2013fast} can be viewed as approximate variational inference schemes \cite{gal2015bayesian, kingma2015variational}. We compare our results to \cite{gal2015bayesian}. Furthermore, structured variational approximations \cite{louizos2017multiplicative}, auxiliary variables \cite{maaloe2016auxiliary}, and Stochastic Gradient MCMC \cite{li2016learning} have been proposed to approximate the intractable posterior probability distribution. Recently, the uncertainty afforded by \textit{Bayes by Backprop} trained neural networks has been used successfully to train feedforward neural networks in both supervised and reinforcement learning environments \cite{blundell2015weight,lipton2016efficient,houthooft2016curiosity}, for training recurrent neural networks \cite{fortunato2017bayesian}, and convolutional neural networks \cite{shridhar2018BayesianComprehensive, neklyudov2018variance}.
\subsection{Bayes by Backprop} \label{BBB}
\textit{Bayes by Backprop} \cite{graves2011practical, blundell2015weight} is a variational inference method to learn the posterior distribution on the weights $w \sim q_{\theta}(w|\mathcal{D})$ of a neural network from which weights $w$ can be sampled in backpropagation. 
Since the true posterior is typically intractable, an approximate distribution $q_{\theta}(w|\mathcal{D})$ is defined that is aimed to be as similar as possible to the true posterior $p(w|\mathcal{D})$, measured by the Kullback-Leibler (KL) divergence \cite{kullback1951information}. Hence, we define the optimal parameters $\theta^{opt}$ as
\begin{equation}
    \begin{aligned} \label{KL}
        \theta^{opt}&=\underset{\theta}{\text{arg min}}\ \text{KL} \ [q_{\theta}(w|\mathcal{D})\|p(w|\mathcal{D})] \\
        &=\underset{\theta}{\text{arg min}}\ \text{KL} \ [q_{\theta}(w|\mathcal{D})\|p(w)]-\mathbb{E}_{q(w|\theta)}[\log p(\mathcal{D}|w)]+\log p(\mathcal{D})
    \end{aligned}
\end{equation}

where
\begin{equation}
    \text{KL} \ [q_{\theta}(w|\mathcal{D})\|p(w)]= \int q_{\theta}(w|\mathcal{D})\log\frac{q_{\theta}(w|\mathcal{D})}{p(w)}dw .
\end{equation}
This derivation forms an optimization problem with a resulting cost function widely known as \textit{variational free energy} \cite{neal1998view,yedidia2005constructing,friston2007variational} which is built upon two terms: the former, $\text{KL} \ [q_{\theta}(w|\mathcal{D})\|p(w)]$, is dependent on the definition of the prior $p(w)$, thus called complexity cost, whereas the latter, $\mathbb{E}_{q(w|\theta)}[\log p(\mathcal{D}|w)]$, is dependent on the data $p(\mathcal{D}|w)$, thus called likelihood cost. 
The term $\log p(\mathcal{D})$ can be omitted in the optimization because it is constant.
\newline Since the KL-divergence is also intractable to compute exactly, we follow a stochastic variational method \cite{graves2011practical,blundell2015weight}.
We sample the weights $w$ from the variational distribution $q_{\theta}(w|\mathcal{D})$ since it is much more probable to draw samples which are appropriate for numerical methods from the variational posterior $q_{\theta}(w|\mathcal{D})$ than from the true posterior $p(w|\mathcal{D})$. Consequently, we arrive at the tractable cost function \eqref{cost} which is aimed to be optimized, i.e. minimised with respect to $\theta$, during training:
\begin{equation} \label{cost}
    \mathcal{F}(\mathcal{D}, \theta)\approx \sum_{i=1}^n \log q_{\theta}(w^{(i)}|\mathcal{D})-\log p(w^{(i)})-\log p(\mathcal{D}|w^{(i)})
\end{equation}
where $n$ is the number of draws. We sample $w^{(i)}$ from $q_{\theta}(w|\mathcal{D})$. 
\subsection{Bayes by Backprop for convolutional neural networks}
In this section, we explain our algorithm of building \acp{cnn} with probability distributions over weights in each filter, as seen in Figure \ref{fig:filter_scalar}, and apply \textit{Bayes by Backprop} to compute the intractable posterior probability distributions $p(w|\mathcal{D})$, as described in the previous section \ref{BBB}. Notably, a fully Bayesian perspective on a \ac{cnn} is for most \ac{cnn} architectures not accomplished by merely placing probability distributions over weights in convolutional layers; it also requires probability distributions over weights in fully-connected layers. 

\subsubsection{Local reparameterization trick for convolutional layers} \label{lrt}
We utilize the local reparameterization trick \cite{kingma2015variational} and apply it to \acp{cnn}. Following \cite{kingma2015variational,neklyudov2018variance}, we do not sample the weights $w$, but we sample layer activations $b$ instead due to its consequent computational acceleration. The variational posterior probability distribution $q_{\theta}(w_{ijhw}|\mathcal{D})=\mathcal{N}(\mu_{ijhw},\alpha_{ijhw}\mu^2_{ijhw})$ (where $i$ and $j$ are the input, respectively output layers, $h$ and $w$ the height, respectively width of any given filter) allows to implement the local reparamerization trick in convolutional layers. This results in the subsequent equation for convolutional layer activations $b$:
\begin{equation}
    b_j=A_i\ast \mu_i+\epsilon_j\odot \sqrt{A^2_i\ast (\alpha_i\odot \mu^2_i)}
\end{equation}
where $\epsilon_j \sim \mathcal{N}(0,1)$, $A_i$ is the receptive field, $\ast$ signalises the convolutional operation, and $\odot$ the component-wise multiplication.

\subsubsection{Applying two sequential convolutional operations (mean and variance)}
The crux of equipping a \ac{cnn} with probability distributions over weights instead of single point-estimates and being able to update the variational posterior probability distribution $q_{\theta}(w|\mathcal{D})$ by backpropagation lies in applying \textit{two} convolutional operations whereas filters with single point-estimates apply \textit{one}. As explained in the previous section \ref{lrt}, we deploy the local reparametrization trick and sample from the activations $b$. Since activations $b$ are functions of mean $\mu_{ijwh}$ and variance $\alpha_{ijhw}\mu^2_{ijhw}$ among others, we are able to compute the two variables determining a Gaussian probability distribution, mean $\mu_{ijhw}$ and variance $\alpha_{ijhw}\mu^2_{ijhw}$, separately. 
\newline We pursue this in two convolutional operations: in the first, we treat the output $b$ as an output of a \ac{cnn} updated by frequentist inference. We optimize with Adam \cite{kingma2014adam} towards a single point-estimate which makes the accuracy of classifications in the validation dataset increasing. We interpret this single point-estimate as the mean $\mu_{ijwh}$ of the variational posterior probability distributions $q_{\theta}(w|\mathcal{D})$. In the second convolutional operation, we learn the variance $\alpha_{ijhw}\mu^2_{ijhw}$. As this formulation of the variance includes the mean $\mu_{ijwh}$, only $\alpha_{ijhw}$ needs to be learned in the second convolutional operation \cite{molchanov2017variational}. In this way, we ensure that only one parameter is updated per convolutional operation, exactly how it would have been with a \ac{cnn} updated by frequentist inference. 
\newline In other words, while we learn in the first convolutional operation the MAP of the variational posterior probability distribution $q_{\theta}(w|\mathcal{D})$, we observe in the second convolutional operation how much values for weights $w$ deviate from this MAP. This procedure is repeated in the fully-connected layers. In addition, to accelerate computation, to ensure a positive non-zero variance $\alpha_{ijhw}\mu^2_{ijhw}$, and to enhance accuracy, we learn $\log \alpha_{ijhw}$ and use the \textit{Softplus} activation function as further described in the Experiments section \ref{experiments}.
\section{Uncertainty estimation in Bayesian \acp{cnn}} \label{uncertainty}
In classification tasks, we are interested in the predictive distribution $p_{\mathcal{D}}(y^*|x^*)$, where $x^*$ is an unseen data example and $y^*$ its predicted class. For a Bayesian neural network, this quantity is given by:
\begin{align}
p_{ \mathcal{D}}(y^*|x^*) = \int p_{w}(y^*|x^*) \ p_{\mathcal{D}}(w) \ dw
\end{align}
In \textit{Bayes by Backprop}, Gaussian distributions $q_{\theta}(w|\mathcal{D}) \sim \mathcal{N}(w|\mu, \sigma^2)$, where $\theta = \{ \mu, \sigma \}$ are learned with some dataset $\mathcal{D} = \{ x_{i}, y_{i} \}_{i=1}^{n}$ as we explained previously in \ref{BBB}. Due to the discrete and finite nature of most classification tasks, the predictive distribution is commonly assumed to be a categorical. Incorporating this aspect into the predictive distribution gives us
\begin{equation}
\begin{aligned}
    p_{\mathcal{D}}(y^*|x^*)& = \int \text{Cat}(y^*|f_w(x^*)) \mathcal{N}(w|\mu, \sigma^2) \ dw\\
    &=  \int \prod_{c=1}^{C} f(x_{c}^*|w)^{y_{c}^*} \frac{1}{\sqrt{2\pi \sigma^2}} e^{-\frac{(w - \mu)^2}{2\sigma^2}} \ dw 
\end{aligned}
\end{equation}
where $C$ is the total number of classes and $\sum_c f(x_{c}^*|w) = 1$.
\newline As there is no closed-form solution due to the lack of conjugacy between categorical and Gaussian distributions, we cannot recover this distribution. However, we can construct an unbiased estimator of the expectation by sampling from $q_{\theta}(w|\mathcal{D})$:
\begin{equation}
\mathbb{E}_{q}[p_{\mathcal{D}}(y^*|x^*)] = \int q_{\theta}(w|\mathcal{D}) \ p_w(y|x) \ dw \approx \frac{1}{T}\sum_{t=1}^{T} p_{w_t}(y^*|x^*)
\end{equation}
where $T$ is the predefined number of samples.
This estimator allows us to evaluate the uncertainty of our predictions by the definition of variance, hence called \textit{predictive variance} and denoted as $\text{Var}_q$:
\begin{equation} \label{variance}
    \begin{aligned} 
&\text{Var}_q\big( p(y^*|x^*) \big) = \mathbb{E}_q[y^*y^{*T}] - \mathbb{E}_q[y^*]\mathbb{E}_q[y^*]^T \\ &= \underbrace{\int \Big[ \text{diag} \Big(\mathbb{E}_p[y^*] \Big) - \mathbb{E}_p[y^*] \ \mathbb{E}_p[y^*]^T \Big] \ q_{\theta}(w|\mathcal{D}) \ dw}_\text{aleatoric} \\ &+ \underbrace{\int \Big( \mathbb{E}_p[y^*] - \mathbb{E}_q[y^*] \Big) \ \Big( \mathbb{E}_p[y^*] - \mathbb{E}_q[y^*] \Big)^T q_{\theta}(w|\mathcal{D}) \ dw}_\text{epistemic}
\end{aligned}
\end{equation}
where $\mathbb{E}_p[y^*] = \mathbb{E}_{p(y^*|x^*)}[y^*]$ and $\mathbb{E}_q[y^*] =  \mathbb{E}_{q_\theta(y^*|x^*)}[y^*]$.
\newline It is important to split the uncertainty in form of the predictive variance into aleatoric and epistemic quantities since it allows the modeler to evaluate the room for improvements: while aleatoric uncertainty (also known as statistical uncertainty) is merely a measure for the variation of ("noisy") data, epistemic uncertainty is caused by the model. Hence, a modeler can see whether the quality of the data is low (i.e. high aleatoric uncertainty), or the model itself is the cause for poor performances (i.e. high epistemic uncertainty). The former can be improved by gathering more data, whereas the latter requests to refine the model \cite{der2009aleatory}.
\subsection{Related work}
As shown in \eqref{variance}, the predictive variance can be decomposed into the aleatoric and epistemic uncertainty. \cite{kendall2017uncertainties} and \cite{kwon2018uncertainty} have proposed disparate methods to do so. In this paragraph, we first review these two methods before we explain our own method and how it overcomes deficiencies of the two aforementioned ones, especially the usage of Softmax in the output layer by \cite{kwon2018uncertainty}. 
\newline \cite{kendall2017uncertainties} derived a method how aleatoric and epistemic uncertainties can directly be estimated by constructing a Bayesian neural network with the last layer before activation consisting of mean and variance of logits, denoted $\hat{\mu}_t$ and $\hat{\sigma}_t^2$. In other words, the pre-activated linear output of the neural network has $2K$ dimensions with $K$ being the number of output units, i.e. potential classes. They propose an estimator as
\begin{equation}
    \text{Var}_q\big( p(y^*|x^*) \big) = \frac{1}{T} \sum_{t=1}^T \text{diag}(\hat{\sigma}_t^2) + \frac{1}{T} \sum_t=1^T (\hat{\mu}_t - \Bar{\mu}) \ (\hat{\mu}_t - \Bar{\mu})^T
\end{equation}
where $\Bar{\mu} = \sum_{t=1}^T \hat{\mu}_t / T$. \cite{kwon2018uncertainty} mention the deficiencies of this approach: first, it models the variability of the linear predictors $\hat{\mu}_t$ and $\hat{\sigma}_t^2$ and not the predictive probabilities; second, it ignores the fact that the covariance matrix of a multinomial random variable is a function of the mean vector; and third, the aleatoric uncertainty does not reflect correlations because of the diagonal matrix.
\newline To overcome these deficiencies, \cite{kwon2018uncertainty} propose
\begin{equation} \label{kwonunc}
    \text{Var}_q\big( p(y^*|x^*) \big) = \underbrace{\frac{1}{T} \sum_{t=1}^T \text{diag}(\hat{p}_t)-\hat{p}_t \ \hat{p}_t^T}_\text{aleatoric} + \underbrace{\frac{1}{T}\sum_{t=1}^T (\hat{p}_t - \Bar{p}) (\hat{p}_t - \Bar{p})^T}_\text{epistemic}
\end{equation}
where $\Bar{p} = \frac{1}{T}\sum_{t=1}^T \hat{p}_t$ and $\hat{p}_t = \text{Softmax}\big ( f_{w_{t}}(x^*) \big )$. By doing so, they do not need the pre-activated linear outputs $\hat{\mu}_t$ and $\hat{\sigma}_t^2$. Consequently, they can directly compute the variability of the predictive probability and do not require additional sampling steps. With increasing $T$, \eqref{kwonunc} converges in probability to \eqref{variance}.
\subsection{Softplus normalization} \label{softplus_normalization}
Deploying the Softmax function assumes we classify according to the logistic sigmoid function:
\begin{align} \label{logsig}
    a &= \text{log} \ \frac{p(c=1 | x)}{p(c=2 | x)} = \frac{p(x|c=1) \ p(c=1)}{p(x|c=2) \ p(c=2)} \\
    \sigma(a) &= \frac{1}{1 + \exp{(-a)}} = p(c=1 | x) 
\end{align} 
where the label $c=1$ is assigned to the input $x$ if $\sigma(a) = p(c=1 | x) = p(y=1 | x^*) \geq 0.5$.
Assume we have $C$ classes, this logistic sigmoid function can be generalized to the Softmax function:
\begin{equation}
    p(c | x) = \frac{p(x|c) \ p(c)}{\sum_{c=1}^C p(x | c) \  p(c)}
\end{equation}
Here, we propose a novel method how aleatoric and epistemic uncertainty estimations can be computed without having an additional non-linear transformation by the generalized logistic sigmoid function in the output layer. 
\newline The predictive variance remains in the form as stated in \eqref{kwonunc}, except the subsequent two amendments compared to \cite{kwon2018uncertainty}: we circumvent the exponential term in the Softmax function, hence generate consistency of activation functions in the entire \ac{cnn}. Despite not punishing wrong predictions particularly with the exponential term, we also resolve concerns about robustness by employing not more than one type of non-linear transformations within the same neural network while we preserve the desideratum of having probabilities as outputs. We implement this in two subsequent computations: first, we employ the same activation function in the output layer as we employ in any other hidden layer, namely the Softplus function, see \eqref{softplus}; second, these outputs are normalized by dividing the output of each class by the sum of all outputs:
\begin{equation}
    y^*_{norm} = \frac{y^*_c}{\sum_C y^*_c}
\end{equation}
This procedure can be summarized as a replacement of $\hat{p}_t = \text{Softmax}\big ( f_{w_{t}}(x^*) \big )$ with 
$\hat{p}_t = \text{Softplus}_{\text{n}}\big ( f_{w_{t}}(x^*) \big )$ where $\text{Softplus}_{\text{n}}$ is the aforementioned normalization of the Softplus output. We call this computation Softplus normalization.
Our intuition for this is as follows: we classify according to a categorical distribution as we outlined previously in section \ref{uncertainty}. This can be seen as an approximation of the outputs with one-hot vectors. If those vectors already contain a lot of zeros, the approximation is going to be more accurate. For a Softmax, predicting zeros requires a logit of $-\infty$ which is hard to achieve in practice. For the Softplus normalization, we just need a negative value roughly smaller than $-4$ to get nearly a zero in the output. Consequently, Softplus normalization can easily produce vectors that are in practice zero, whereas Softmax cannot. In sum, our proposed method is as follows:
\begin{equation} \label{softnormunc}
    \text{Var}_q\big( p(y^*|x^*) \big) = \underbrace{\frac{1}{T} \sum_{t=1}^T \text{diag}(\hat{p}_t)-\hat{p}_t \ \hat{p}_t^T}_\text{aleatoric} + \underbrace{\frac{1}{T}\sum_{t=1}^T (\hat{p}_t - \Bar{p}) (\hat{p}_t - \Bar{p})^T}_\text{epistemic}
\end{equation}
where $\Bar{p} = \frac{1}{T}\sum_{t=1}^T \hat{p}_t$ and $\hat{p}_t = \text{Sofplus}_{\text{n}}\big ( f_{w_{t}}(x^*) \big )$.
\section{Experiments} \label{experiments}
For all conducted experiments, we implement the foregoing description of Bayesian \acp{cnn} with variational inference in LeNet-5 \cite{lecun1998gradient}, AlexNet \cite{krizhevsky2012imagenet}, and VGG \cite{simonyan2014very}. The exact architecture specifications can be found in the Appendix \ref{appendix} and in our GitHub repository\footnote{\url{https://github.com/kumar-shridhar/PyTorch-Softplus-Normalization-Uncertainty-Estimation-Bayesian-CNN}}.
We train the networks with the MNIST dataset of handwritten digits \cite{lecun1998gradient}, and with the CIFAR-10 and CIFAR-100 datasets \cite{krizhevsky2009learning} since these datasets serve widely as benchmarks for \acp{cnn}' performances. The originally chosen activation functions in all architectures are ReLU, but we must introduce another, called Softplus, see \eqref{softplus}, because of our method to apply two convolutional or fully-connected operations. As aforementioned, one of these is determining the mean $\mu$, and the other the variance $\alpha \mu^2$. Specifically, we apply the Softplus function because we want to ensure that the variance $\alpha \mu^2$ never becomes zero. This would be equivalent to merely calculating the MAP, which can be interpreted as equivalent to a maximum likelihood estimation (MLE), which is further equivalent to utilising single point-estimates, hence frequentist inference. The Softplus activation function is a smooth approximation of ReLU. Although it is practically not influential, it has the subtle and analytically important advantage that it never becomes zero for $x \rightarrow -\infty$, whereas ReLU becomes zero for $x \rightarrow -\infty$.
\begin{equation}\label{softplus}
     \text{Softplus}(x) = \frac{1}{\beta} \cdot \log \big ( 1 + \exp(\beta \cdot x) \big )
\end{equation}
where $\beta$ is by default set to $1$.
\newline All experiments are performed with the same hyper-parameters settings as stated in the Appendix \ref{appendix}.

\begin{comment}
\subsection{Datasets}
As aforementioned, we train various architectures on multiple datasets, namely MNIST, CIFAR-10, and CIFAR-100. 
\newline
\textbf{Classification on MNIST.}
The MNIST dataset of handwritten digits consists of 60,000 training and 10,000 validation images of 28 by 28 pixels. Each image is labelled with its corresponding number (between zero and nine, inclusive).
\newline
\textbf{Classification on CIFAR-10.}
The CIFAR-10 dataset consists of 60,000 colour images in 10 classes, with 6,000 images per class, each image 32 by 32 pixels large. Each of the classes has 5,000 training images and 1,000 validation images. 
\newline
\textbf{Classification on CIFAR-100.}
This dataset is similar to the CIFAR-10, except it has 100 classes containing 600 images each. There are 500 training images and 100 validation images per class. The resolution of the images is as in CIFAR-10 32 by 32 pixels.
\end{comment}

\subsection{Results of Bayesian CNNs with variational inference}
\begin{figure}[t] 
\centering
\includegraphics[width=0.5\linewidth]{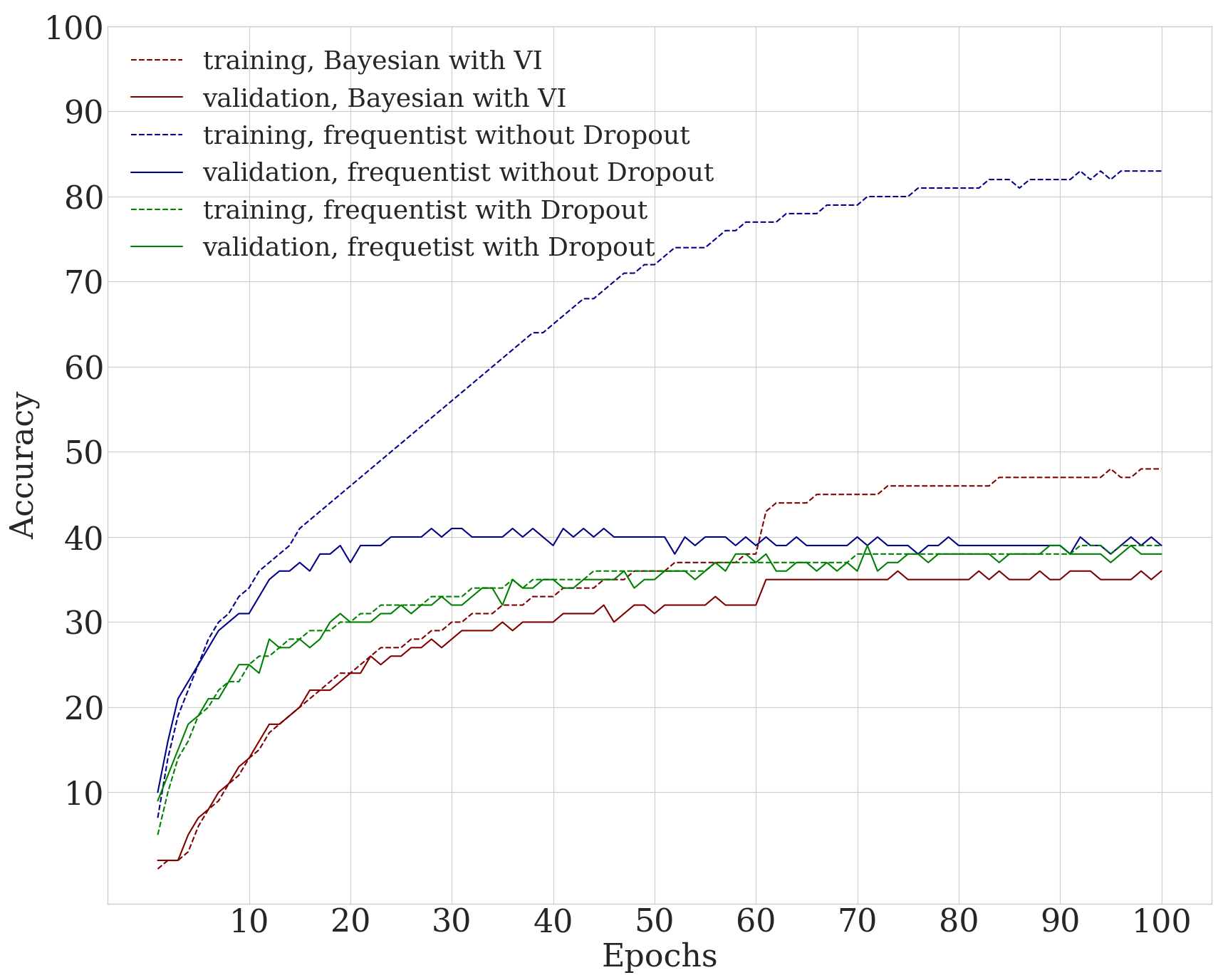}
\caption{AlexNet trained on CIFAR-100 by Bayesian and frequentist inference. The frequentist AlexNet without Dropout overfits while the Bayesian AlexNet naturally incorporates an effect of regularization, comparable to a frequentist AlexNet with three Dropout layers.}
\label{fig:regularisation}
\end{figure}

\begin{table}[b]
\tiny
    \centering
    \renewcommand{\arraystretch}{1.5}
    \resizebox{0.5\linewidth}{!}{
    \begin{tabular}{ l  c  c  c  c } 
     \hline
      \empty & MNIST & CIFAR-10 & CIFAR-100  \\ [0.75ex]
     \hline
     Bayesian VGG (with VI) & 99 & 86 & 45 \\
     Frequentist VGG & 99 & 85 & 48 \\
     \hdashline
     
     Bayesian AlexNet (with VI) & 99 & 73 & 36  \\
     
     Frequentist AlexNet & 99 & 73 & 38  \\
     \hdashline
     Bayesian LeNet-5 (with VI) & 98 & 69 & 31  \\
     
     Frequentist LeNet-5 & 98 & 68 & 33  \\
     \hdashline
     Bayesian LeNet-5 (with Dropout) & 99 & 83 & \empty \\ 
     \hline \\
    \end{tabular}} 
    \renewcommand{\arraystretch}{1.5}
    \caption{Comparison of validation accuracies (in percentage) for different architectures with variational inference (VI), frequentist inference and Dropout as a Bayesian approximation as proposed by Gal and Ghahramani \cite{gal2015bayesian} for MNIST, CIFAR-10, and CIFAR-100.}
    \label{tab:results}
\end{table}
We evaluate the performance of our Bayesian \acp{cnn} with variational inference. Table \ref{tab:results} shows a comparison of validation accuracies (in percentage) for architectures trained by two disparate Bayesian approaches, namely variational inference, i.e. \textit{Bayes by Backprop} and Dropout as proposed by Gal and Ghahramani \cite{gal2015bayesian}, plus frequentist inference for all three datasets. Bayesian \acp{cnn} trained by variational inference achieve validation accuracies comparable to their counter-architectures trained by frequentist inference. On MNIST, validation accuracies of the two disparate Bayesian approaches are comparable, but a Bayesian LeNet-5 with Dropout achieves a considerable higher validation accuracy on CIFAR-10, although we were not able to reproduce these reported results.
\newline In Figure \ref{fig:regularisation}, we show how Bayesian networks incorporate naturally effects of regularization, exemplified on AlexNet. While an AlexNet trained by frequentist inference without any regularization overfits greatly on CIFAR-100, an AlexNet trained by Bayesian inference on CIFAR-100 does not. It performs equivalently to an AlexNet trained by frequentist inference with three layers of Dropout after the first, fourth, and sixth layers in the architecture. In initial epochs, Bayesian \acp{cnn} trained by variational inference start with a low validation accuracy compared to architectures trained by frequentist inference. Initialization for both inference methods is chosen equivalently: the variational posterior probability distributions $q_{\theta}(w|\mathcal{D})$ is initially approximated as standard Gaussian distributions, while initial point-estimates in architectures trained by frequentist inference are randomly drawn from a standard Gaussian distribution. These initialization methods ensure weights are neither too small nor too large in the beginning of training.
\begin{comment}
%
\begin{figure}[b]
\centering
\begin{minipage}{0.35\textwidth}
\includegraphics[width=\linewidth]{distribution.png}
\end{minipage}
%
\begin{minipage}{0.29\textwidth}
\includegraphics[width=\linewidth]{std_CNN.png}
\end{minipage}
\caption{Convergence of the Gaussian variational posterior probability distribution $q_{\theta}(w|\mathcal{D})$ (left) and its standard deviation (right) of a random model parameter at epochs 1, 5, 20, 50, and 100. CIFAR-10 is trained on Bayesian LeNet-5.}
\label{fig:distributions}
\end{figure} 
%
\newline Figure \ref{fig:distributions} displays the convergence of the standard deviation $\sigma$ of the variational posterior probability distribution $q_{\theta}(w|\mathcal{D})$ of a random model parameter over epochs. All prior probability distributions $p(w)$ are initialized as standard Gaussians. The variational posterior probability distributions $q_{\theta}(w|\mathcal{D})$ are approximated as Gaussian distributions which become more confident as more data is processed - observable by the decreasing standard deviation over epochs in Figure \ref{fig:distributions}. Although the validation accuracy for MNIST on Bayesian LeNet-5 has already converged at 99\%, we can still see a considerably steep decrease in the parameter's standard deviation. In Figure \ref{fig:distributions}, we plot the actual Gaussian variational posterior probability distributions $q_{\theta}(w|\mathcal{D})$ of a random parameter of LeNet-5 trained on CIFAR-10 at some epochs.
\end{comment}
%
\subsection{Results of uncertainty estimations by Softplus normalization}
\begin{figure}[b] 
\centering
\begin{minipage}{0.15\textwidth}
\centering
\text{$\gamma = $}\par\medskip
\text{$\text{aleatoric} = $}\par\medskip
\includegraphics[width=\linewidth]{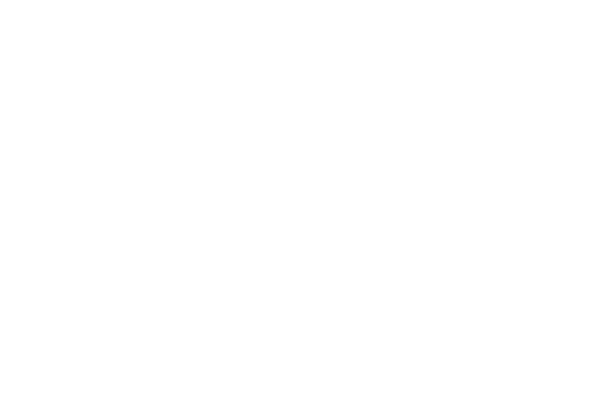}
\end{minipage}
\begin{minipage}{0.15\textwidth}
\centering
\text{$0$}\par\medskip
\text{$0.00099897$}\par\medskip
\includegraphics[width=\linewidth]{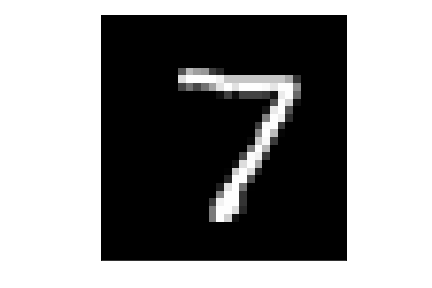}
\end{minipage}
\begin{minipage}{0.15\textwidth}
\centering
\text{$0.1$}\par\medskip
\text{$0.00099885$}\par\medskip
\includegraphics[width=\linewidth]{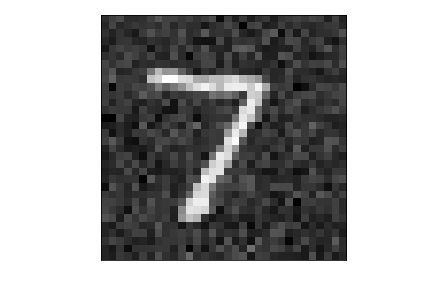}
\end{minipage}
\begin{minipage}{0.15\textwidth}
\centering
\text{$0.2$}\par\medskip
\text{$0.00099875$}\par\medskip
\includegraphics[width=\linewidth]{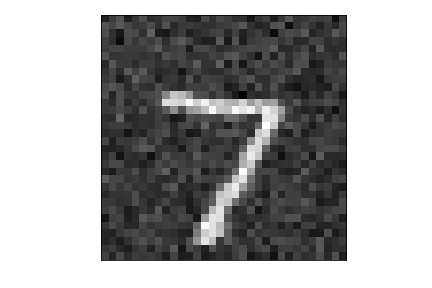}
\end{minipage}
\begin{minipage}{0.15\textwidth}
\centering
\text{$0.3$}\par\medskip
\text{$0.00099838$}\par\medskip
\includegraphics[width=\linewidth]{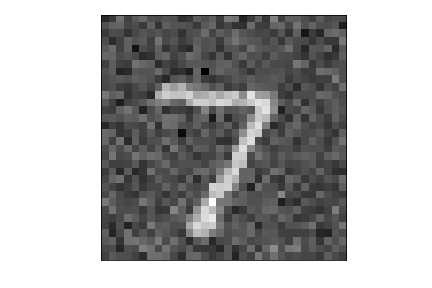}
\end{minipage}
\caption{Aleatoric uncertainty is computed with Bayesian VGG on the same MNIST input image with different noise levels $\gamma$.}
\label{noiseimages}
\end{figure}

Our results for the Softplus normalization estimation method of aleatoric and epistemic uncertainties are combined in Table \ref{tab:uncertainty}. It compares the means over epochs of aleatoric and epistemic uncertainties for our Bayesian \acp{cnn} LeNet-5, AlexNet and VGG. We train on MNIST and CIFAR-10, and compute homoscedastic uncertainties by averaging over all classes, i.e. over (heteroscedastic) uncertainties of each class. We see a correlating pattern between validation accuracies and epistemic uncertainty: with increasing validation accuracy, epistemic uncertainty decreases, observable by the different models. In contrast, aleatoric uncertainty measures the irreducible variability of the datasets, hence is only dependent on the datasets and not the models, which can be seen by the constant aleatoric uncertainties of each dataset across models. 
\begin{table}[t]
\tiny
    \centering
    \renewcommand{\arraystretch}{1.5}
    \resizebox{0.7\linewidth}{!}{
    \begin{tabular}{ l  c  c |  c  } 
     \hline
      \empty & Aleatoric uncertainty &  Epistemic uncertainty & Validation accuracy  \\ [0.75ex]
     \hline
     Bayesian VGG (MNIST) & 0.00110 & 0.0004 & 99 \\
     
     Bayesian VGG (CIFAR-10) & 0.00099 & 0.0013 & 85  \\
     \hdashline
     Bayesian AlexNet (MNIST) & 0.00110 & 0.0019 & 99 \\
     
     Bayesian AlexNet (CIFAR-10) & 0.00099 & 0.0002 & 73 \\
     \hdashline
     Bayesian LeNet-5 (MNIST) & 0.00110 & 0.0026 & 98   \\
     
     Bayesian LeNet-5 (CIFAR-10) & 0.00099 & 0.0404 & 69   \\
     \hline \\
     
    \end{tabular}}
    \renewcommand{\arraystretch}{1.5}
    \caption{Aleatoric and epistemic uncertainty for Bayesian VGG, AlexNet and LeNet-5 calculated for MNIST and CIFAR-10, computed by Softplus normalization \eqref{softplus_normalization}. Validation accuracy is displayed to demonstrate the negative correlation between validation accuracy and epistemic uncertainty.}
    \label{tab:uncertainty}
\end{table}
\newline We further investigate influences on aleatoric uncertainty estimations by additive standard Gaussian noise of different levels. Here, a random sample of the standard Gaussian distribution is multiplied by a level $\gamma$ and added to each pixel value. We test for $\gamma = 0, 0.1, 0.2, 0.3$ and see that aleatoric uncertainty is independent of the added noise level (see Figure \ref{noiseimages}), i.e. aleatoric uncertainty is constant to six decimal places. Our intuition for this phenomenon is as follows: by normalizing the output of the Softplus function among one batch, of which all images have the same added level of noise, the aleatoric uncertainty captures the variability among the images in \textit{one} batch - and not across batches with different levels of noise. 
\section{Conclusion}
We propose a novel method how aleatoric and epistemic uncertainties can be estimated in Bayesian \acp{cnn}. We call this method Softplus normalization. Firstly, we discuss briefly the differences of a \textit{Bayes by Backprop} based variational inference to the Dropout based approximation of Gal \& Ghahramani \cite{gal2015bayesian}. Secondly, we derive the Softplus normalization uncertainty estimation method with previously published methods for these uncertainties. We evaluate our approach on three datasets (MNIST, CIFAR-10, CIFAR-100).
\newline As a base, we show that Bayesian \acp{cnn} with variational inference achieve comparable results as those achieved by the same network architectures trained by frequentist inference, but include naturally a regularization effect and an uncertainty measure. Building on that, we examine how our proposed Softplus normalization method to estimate aleatoric and epistemic uncertainties is derived by previous work in this field, how it differs from those, and explain why it is more appropriate to use in computer vision than these previously proposed methods.

\small
\bibliographystyle{unsrt}
\bibliography{neurips_2019}

\newpage
\onecolumn
\section{Appendix}\label{appendix}
\subsection{Experiment specifications}
\begin{table}[h]
\tiny
    \centering
    \tiny
    \renewcommand{\arraystretch}{2}
    \begin{tabular}[c]{c | c} 
     \hline
     variable & value \\ [0.5ex] 
     \hline
     learning rate &  0.001\\ 
     
     epochs & 100 \\
     
     batch size & 128 \\
     
     sample size & 10 \\
     
     $(\alpha \mu^2)_{init}$ of approximate posterior $q_{\theta}(w|\mathcal{D})$ & -10 \\
     
     optimizer & Adam \cite{kingma2014adam} \\
     
     $\lambda$ in $\ell$-2 normalization & 0.0005 \\
    
     $\beta_i$ & $\frac{2^{M-i}}{2^M-1}$ \cite{blundell2015weight} \\ [1ex] 
     \hline
    \end{tabular} 
    \renewcommand{\arraystretch}{2}
\end{table}
\subsection{Model architectures}
\subsubsection{LeNet-5}
\begin{table}[h!]
    \centering
    \tiny
    \renewcommand{\arraystretch}{2}
    \begin{tabular}{c c c c c c} 
     \hline
     layer type & width & stride & padding & input shape & nonlinearity \\ [0.5ex] 
     \hline
     convolution ($5\times5$) & 6 & 1 & 0 & $M\times1\times32\times32$ & Softplus \\ 
     
     Mmax-pooling ($2\times2$) & \empty & 2 & 0 & $M\times6\times28\times28$ & \empty \\
     
     convolution ($5\times5$) & 16 & 1 & 0 & $M\times1\times14\times14$ & Softplus \\
     
     max-pooling ($2\times2$) & \empty & 2 & 0 & $M\times16\times10\times10$ & \empty \\
    
     fully-connected & 120 & \empty & \empty & $M\times400$ & Softplus \\
     
     fully-connected & 84 & \empty & \empty & $M\times120$ & Softplus \\
     
     fully-connected & 10 & \empty & \empty & $M\times84$ & Softplus normalization \\ [1ex] 
     \hline
    \end{tabular} 
    \renewcommand{\arraystretch}{1}
    \label{tab:LeNet}
\end{table}

\subsubsection{AlexNet}
\begin{table}[h!]
    \centering
    \tiny
    \renewcommand{\arraystretch}{2}
    \begin{tabular}{c c c c c c} 
 \hline
 layer type & width & stride & padding & input shape & nonlinearity \\ [0.5ex] 
 \hline
 convolution ($11\times11$) & 64 & 4 & 5 & $M\times3\times32\times32$ & Softplus \\ 
 
 max-pooling ($2\times2$) & \empty & 2 & 0 & $M\times64\times32\times32$ & \empty \\
 
 convolution ($5\times5$) & 192 & 1 & 2 & $M\times64\times15\times15$ & Softplus \\
 
 max-pooling ($2\times2$) & \empty & 2 & 0 & $M\times192\times15\times15$ & \empty \\
 
 convolution ($3\times3$) & 384 & 1 & 1 & $M\times192\times7\times7$ & Softplus \\
 
 convolution ($3\times3$) & 256 & 1 & 1 & $M\times384\times7\times7$ & Softplus \\
 
 convolution ($3\times3$) & 128 & 1 & 1 & $M\times256\times7\times7$ & Softplus \\
 
 max-pooling ($2\times2$) & \empty & 2 & 0 & $M\times128\times7\times7$ & \empty \\
 
 fully-connected & 128 & \empty & \empty & $M\times128$ & Softplus normalization \\ [1ex] 
 \hline
\end{tabular}
\renewcommand{\arraystretch}{1}
\label{tab:AlexNet}
\end{table}

\newpage
\subsubsection{VGG}
\begin{table}[h!]
    \centering
    \tiny
    \renewcommand{\arraystretch}{2}
    \begin{tabular}{c c c c c c} 
 \hline
 layer type & width & stride & padding & input shape & nonlinearity \\ [0.5ex] 
 \hline
 convolution ($3\times3$) & 64 & 1 & 1 & $M\times3\times32\times32$ & Softplus \\ 
 
 convolution ($3\times3$) & 64 & 1 & 1 & $M\times64\times32\times32$ & Softplus \\
 
 max-pooling ($2\times2$) & \empty & 2 & 0 & $M\times64\times32\times32$ & \empty \\
 
 convolution ($3\times3$) & 128 & 1 & 1 & $M\times64\times16\times16$ & Softplus \\
 
 convolution ($3\times3$) & 128 & 1 & 1 & $M\times128\times16\times16$ & Softplus \\
 
 max-pooling ($2\times2$) & \empty & 2 & 0 & $M\times128\times16\times16$ & \empty \\
 
 convolution ($3\times3$) & 256 & 1 & 1 & $M\times128\times8\times8$ & Softplus \\
  
 convolution ($3\times3$) & 256 & 1 & 1 & $M\times256\times8\times8$ & Softplus \\
 
 convolution ($3\times3$) & 256 & 1 & 1 & $M\times256\times8\times8$ & Softplus \\
 
 max-pooling ($2\times2$) & \empty & 2 & 0 & $M\times256\times8\times8$ & \empty \\
 
 convolution ($3\times3$) & 512 & 1 & 1 & $M\times256\times4\times4$ & Softplus \\
 
 convolution ($3\times3$) & 512 & 1 & 1 & $M\times512\times4\times4$ & Softplus \\
 
 convolution ($3\times3$) & 512 & 1 & 1 & $M\times512\times4\times4$ & Softplus \\
 
 max-pooling ($2\times2$) & \empty & 2 & 0 & $M\times512\times4\times4$ & \empty \\
 
 convolution ($3\times3$) & 512 & 1 & 1 & $M\times512\times2\times2$ & Softplus \\
 
 convolution ($3\times3$) & 512 & 1 & 1 & $M\times512\times2\times2$ & Softplus \\
 
 convolution ($3\times3$) & 512 & 1 & 1 & $M\times512\times2\times2$ & Softplus \\
 
 max-pooling ($2\times2$) & \empty & 2 & 0 & $M\times512\times2\times2$ & \empty \\

 fully-connected & 512 & \empty & \empty & $M\times512$ & Softplus normalization \\ [1ex] 
 \hline
\end{tabular}
\renewcommand{\arraystretch}{1}
\label{tab:VGG}
\end{table}
\end{document}